\documentclass[twocolumn]{svjour3}          % twocolumn

\usepackage{graphicx}  % Written by David Carlisle and Sebastian Rahtz
\usepackage{subfigure} % Written by Steven Douglas Cochran
\usepackage{natbib}
\usepackage{amsmath}   % From the American Mathematical Society
\usepackage{multirow}
\usepackage{times}
\usepackage{amssymb}
\usepackage[ruled, vlined]{algorithm2e}
\usepackage{algorithmic}

\smartqed  % flush right qed marks, e.g. at end of proof
\journalname{IJCV}

\begin{document}

\title{Transferrable Feature and Projection Learning with Class Hierarchy for Zero-Shot Learning
%\thanks{Grants or other notes about the article }
}



\author{Aoxue Li \and Zhiwu Lu \and Jiechao Guan \and Tao Xiang \and Liwei~Wang \and Ji-Rong Wen}


\institute{
A. Li and L. Wang \at
The Key Laboratory of Machine Perception (MOE), School of Electronics Engineering and Computer Science, Peking University, Beijing 100871, China\\
\email{lax@pku.edu.cn} \and
Z. Lu, J. Guan and J. Wen \at
The Beijing Key Laboratory of Big Data Management and Analysis Methods, School of Information, Renmin University of China, Beijing 100872, China \\
\email{zhiwu.lu@gmail.com} \and
T. Xiang \at
The School of Electronic Engineering and Computer Science, Queen Mary University of London, Mile End Road, London E1 4NS, United Kingdom\\
\email{t.xiang@qmul.ac.uk.}}

\date{Received: date / Accepted: date}

\maketitle

\begin{abstract}
Zero-shot learning (ZSL) aims to transfer knowledge from seen classes to unseen ones so that the latter can be recognised without any training samples. This is made possible by learning a projection function between a feature space and a semantic space (e.g. attribute space). Considering the seen and unseen classes as two domains, a big domain gap often exists which challenges ZSL. Inspired by the fact that an unseen class is not exactly `unseen' if it belongs to the same superclass as a seen class, we propose a novel inductive ZSL model that leverages superclasses as the bridge between seen and unseen classes to narrow the domain gap. Specifically, we first build a class hierarchy of multiple superclass layers and a single class layer, where the superclasses are automatically generated by data-driven clustering over the semantic representations of all seen and unseen class names. We then exploit the superclasses from the class hierarchy to tackle the domain gap challenge in two aspects: deep feature learning and projection function learning. First, to narrow the domain gap in the feature space, we integrate a recurrent neural network (RNN) defined with the superclasses into a convolutional neural network (CNN), in order to enforce the superclass hierarchy. Second, to further learn a transferrable projection function for ZSL, a novel projection function learning method is proposed by exploiting the superclasses to align the two domains. Importantly, our transferrable feature and projection learning methods can be easily extended to a closely related task -- few-shot learning (FSL). Extensive experiments show that the proposed model significantly outperforms the state-of-the-art alternatives in both ZSL and FSL tasks.
\end{abstract}

% main text
\section{Introduction}

\begin{figure*}[t]
\vspace{0.02in}
\begin{center}
\includegraphics[width=0.88\textwidth]{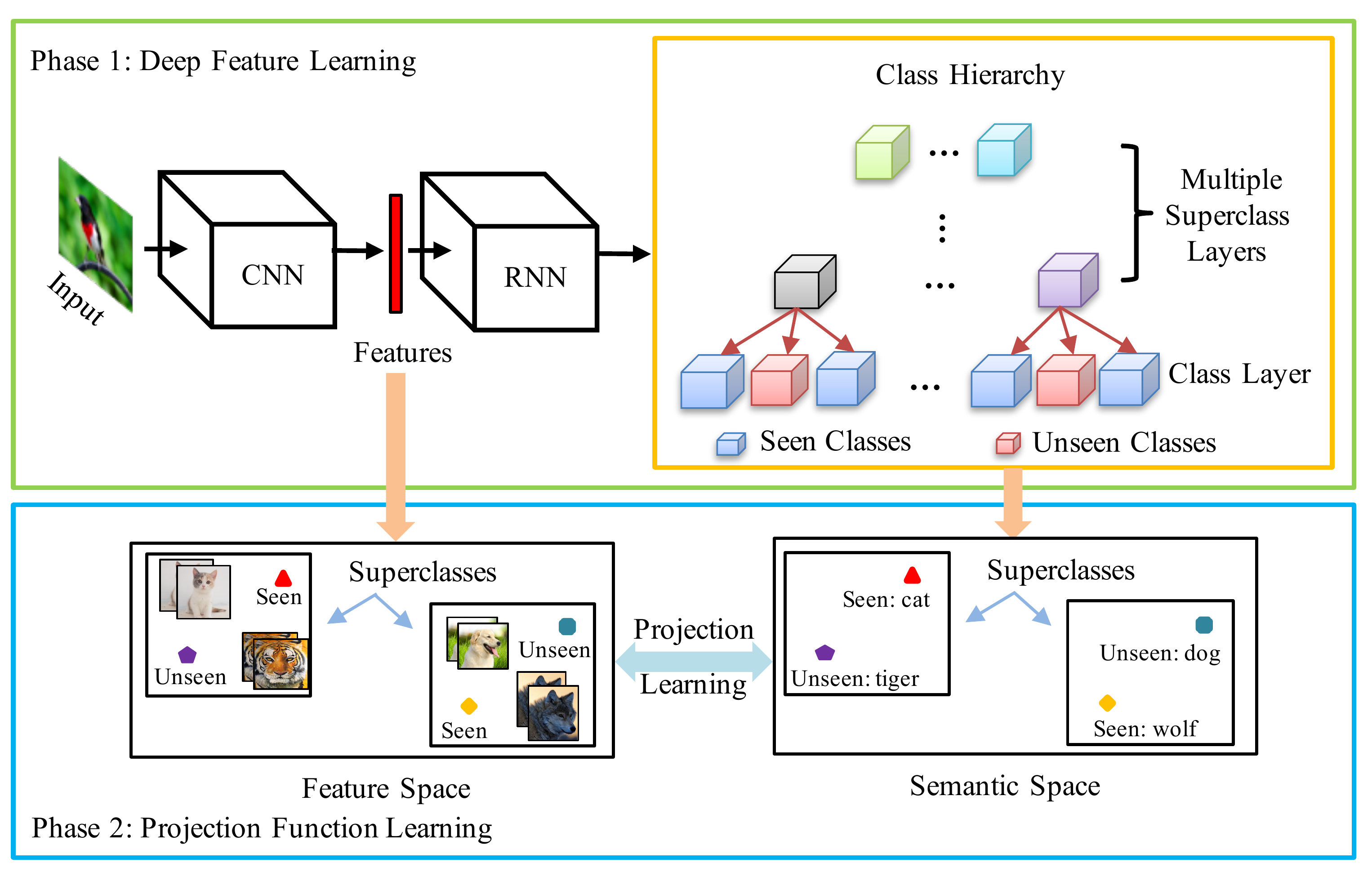}
\end{center}
\vspace{-0.2in}
\caption{Overview of the proposed model for inductive ZSL with class hierarchy (including deep feature learning and projection function learning). }
\label{Fig.1}
\vspace{0.02in}
\end{figure*}

In the past five years, deep neural network (DNN) based models \citep{Gao2017cvpr,donahue2014decaf} have achieved super-human performance on the ILSVRC 1K recognition task. However, most existing object recognition models, particularly those DNN-based ones, require hundreds of image samples to be collected for each object class; many of the object classes are rare and it is thus extremely hard, sometimes impossible to collect sufficient training samples, even with social media. One way to bypass the difficulty in collecting sufficient training data for object recognition is zero-shot learning (ZSL) \citep{rahman2017unified,li2017tgrs,zhang2016eccv,chao2016empirical}. The goal of ZSL is to recognise a new/unseen class without any training samples from the class. A ZSL model typically assumes that each class name is embedded in a semantic space. With this semantic space and a visual feature space representing the appearance of an object in an image, it chooses a joint embedding space (which is mostly formed with the semantic space) and learns a projection function so that both the visual features and the semantic vectors are embedded in the same space. Under the inductive ZSL setting, the projection function is learned only with the seen class training samples and then directly used to project the unseen class samples to the joint embedding space. The class label of a test unseen class sample is assigned to the nearest unseen class prototype.

One of the biggest challenges in ZSL is the domain gap between the seen and unseen classes. As mentioned above, the projection functions learned from the seen classes are applied to the unseen class data. However, the unseen classes are often visually very different from the seen ones; therefore the domain gap between the seen and unseen class domains can be big, meaning that the same projection function may not be able to project an unseen class image to be close to its corresponding class name in the joint embedding space for correct recognition.

To tackle the domain gap challenge, most previous works focus on learning more transferrable projection functions. Specifically, many ZSL models resort to transductive learning \citep{ye2017cvpr,guo2016aaai,Kodirov2015ICCV}, where the unlabeled test samples from unseen classes are used to learn the projection functions. However, such an approach assumes the access to the whole test dataset beforehand, and thus deviates from the motivation of ZSL: the unseen class samples are rare. These transductive ZSL models are thus limited in their practicality in real-world scenarios. Moreover, other than projection function learning, deep feature learning has been largely overlooked in ZSL. Most previous works, if not all, simply use visual features extracted by convolutional neural network (CNN) models pretrained on ImageNet \citep{Russakovsky2015ImageNet}. We argue that deep feature learning and projection function learning should be equally important for overcoming the domain gap challenge.

In this paper, we propose a novel inductive ZSL model that leverages the superclasses as the bridge between seen and unseen classes to narrow the domain gap. The idea is simple: an unseen class is not strictly `zero-shot' if it falls into a superclass that contains one or more seen classes. Exploiting the shared superclass among seen and unseen classes thus provides a means for narrowing the domain gap. In this work, we generate the superclasses by a data-driven approach, without the need of a human-annotated taxonomy. Specifically, we construct a tree-structured class hierarchy that consists of multiple superclass layers and a single class layer. In this tree-structured hierarchy, we take both seen classes and unseen classes as the leaves, and generate the superclasses by clustering over the semantic representations of all seen and unseen class names (see the orange box in Fig.~\ref{Fig.1}). Moreover, by exploiting the superclasses from the class hierarchy, the proposed model aims to narrow the domain gap in two aspects: deep feature learning and projection function learning.

To learn transferrable visual features, we propose a novel deep feature learning model based on the superclasses from the class hierarchy. Specifically, a CNN model is first designed for both class and superclass classification. Since the seen and unseen classes may have the same superclasses, directly training this CNN model enables us to learn transferrable deep features. To further strengthen the feature transferability, we explicitly encode the hierarchical structure among classes/superclasses into the CNN model, by plugging recurrent neural network (RNN) \citep{lstm,zheng2015crf_rnn,liuCVPR17} components into the network. Overall, a novel CNN-RNN architecture is designed for transferrable deep feature learning, as illustrated in the green box in Fig.~\ref{Fig.1}.

To learn a transferrable projection function for inductive ZSL, we propose a novel projection function learning method by utilising the superclasses to align the two domains, as shown in the blue box in Fig.~\ref{Fig.1}. Specifically, we formulate the projection function learning on each superclass layer as a graph regularised self-reconstruction problem. An efficient iterative algorithm is developed as the solver. The results of multiple superclass layers are combined to boost the performance of projection function learning on the single class layer.

Importantly, our model can be easily extended to a closely related problem -- few-shot learning (FSL) \citep{Snell2017nips,Qiao2018cvpr,Finn2017icml,Sung2018cvpr}. Specifically, the above transferrable feature and projection learning methods involved in our model are employed to solve the FSL problem without any modification. Experiments on the widely-used mini-ImageNet dataset \citep{Snell2017nips} show that our model achieves the state-of-the-art results. This suggests that the class hierarchy is also important for narrowing down the domain gap in the FSL problem.

Our contributions are: (1) A novel inductive ZSL model is proposed to align the seen and unseen class domains by utilising the superclasses shared across the two domains. To our best knowledge, this is the first time that the superclasses generated by data-driven clustering have been leveraged in both feature learning and projection learning to narrow the domain gap for inductive ZSL. (2) Due to the domain alignment using the superclasses from the class hierarchy, we have created the new state-of-the-art for ZSL and FSL.

\section{Related Work}

\subsection{Semantic Space}

Three semantic spaces are typically used for ZSL: the attribute space \citep{Zhang2015iccv,guo2016aaai,shojaee2016semi}, the word vector space \citep{Frome2013nips,Norouzi14iclr,Fu2015CVPR}, and the textual description space \citep{reed2016learning,Alexiou2016iccp,Fu2016CVPR}. The attribute space is the most widely-used semantic space. However, for large-scale problems, annotating attributes for each class becomes very difficult. Recently, the semantic word vector space has begun to be popular especially in large-scale problems \citep{Fu2015CVPR,Kodirov2015ICCV}, since no manually defined ontology is required and any class name can be represented as a word vector for free. Beyond semantic attribute or word vector, the semantic space can also be created by directly learning from textual descriptions of categories such as Wikipedia articles \citep{Fu2016CVPR} and sentence descriptions \citep{reed2016learning}.

\subsection{Projection Domain Shift}

To overcome the projection domain shift \citep{fu2015transductive} caused by the domain gap in ZSL, transductive ZSL has been proposed to learn the projection function with not only labelled data from seen classes but also unlabelled data from unseen classes. According to whether the predicted labels of the test images are iteratively used for model learning, existing transductive ZSL models can be divided into two groups: (1) The first group of models \citep{fu2015transductive,Fu2015CVPR,rohrbach2013transfer,ye2017cvpr} first constructs a graph in the semantic space and then the domain gap is reduced by label propagation \citep{wang2008label} on the graph. A variant is the structured prediction model \citep{zhang2016eccv} which utilises a Gaussian parameterisation of the unseen class domain label predictions. (2) The second category of models \citep{guo2016aaai,Kodirov2015ICCV,li2015iccv,shojaee2016semi,
wang2017ijcv,yu2017transductive} involve using the predicted labels of the unseen class data in an iterative model update/adaptation  process as in self-training \citep{xu2015icip,xu2017ijcv}. However, these transductive ZSL models assume the access to the whole test dataset, which is often invalid in the context of ZSL because new classes typically appear dynamically and unavailable before model learning. Instead of assuming the access to all test unseen class data for transductive learning, our model is developed based on inductive learning, and it resorts to learning visual features and projection function only with the training labelled seen class data to counter the projection domain shift.

\subsection{ZSL with Superclasses}

In the area of ZSL, little attention has been paid to ZSL with superclasses. There are only two exceptions. One is \citep{Hwang14}, which learns the relation between attributes and superclasses for semantic embedding, and the other is \citep{Akata2015CVPR} which utilises superclasses to define a semantic space for ZSL. However, the superclasses used in these works are obtained from the manually-defined hierarchical taxonomy, which needs additional cost to collect. In this paper, our model is more flexible by generating superclasses automatically with data-driven clustering over semantic representations of all seen/unseen class names. In addition, the superclasses are not induced into the feature learning in \citep{Akata2015CVPR,Hwang14}.

\vspace{-0.05in}
\subsection{ZSL with Feature Learning}

Most existing ZSL models extract visual features by using CNNs pretrained on ImageNet. This is to assume that the feature extraction model will generalise equally well for seen and unseen classes. This assumption is clearly invalid, particularly under the recently proposed `pure' ZSL setting \citep{Xian2017CVPR}, where the unseen classes have no overlapping with the ImageNet 1K classes used to train the feature extraction model. The only notable exception is \citep{long2017zero}  which uses the central loss and fine-tunes ImageNet pretrained deep model on seen classes. However, without any transferrable learning module, the feature extraction model proposed in \citep{long2017zero} has no guarantee of generalising well to unseen classes.

\subsection{Label Correlation Modeling}

Although no existing ZSL work has considered modeling label correlations to address the domain gap issue, the idea of using the relationship of class labels to improve multi-label classification has been exploited \citep{hex,zheng2015crf_rnn,liuCVPR17}. The label relationship has been modeled as  hierarchy and exclusion graph (HEX) \citep{hex}, CRF \citep{zheng2015crf_rnn} or RNN \citep{liuCVPR17}. Our model differs in two aspects: (1) It is designed to learn transferrable deep features and projection functions, rather than making more accurate label prediction; and (2) the label correlation is guided by a class hierarchy rather than exhaustive as in CRF \citep{zheng2015crf_rnn} and defined by the label frequency as in RNN \citep{liuCVPR17}.

\subsection{Few-Shot Learning}
Few-shot learning (FSL) \citep{Snell2017nips, Qiao2018cvpr, Finn2017icml,Sung2018cvpr} aims to recognise novel classes from very few labelled examples. Such label scarcity issue challenges the standard fine-tuning strategy used in deep learning. Data augmentation can alleviate the label scarcity issue, but cannot solve it. The latest FSL approaches thus choose to transform the deep network training process to meta learning where the transferrable knowledge is learned in the form of good initial conditions, embeddings, or optimisation strategies \citep{Finn2017icml,Sung2018cvpr}. In this paper, we directly employ our transferrable feature learning and projection learning methods to solve FSL task. Experimental results in Sec. \ref{sect:res_fsl} demonstrate that, similar to the ZSL task, our transferrable feature learning and projection learning methods can also achieve state-of-the-art results in the FSL task.

\begin{figure*}[t]
\vspace{0.02in}
\begin{center}
\includegraphics[width=0.96\textwidth]{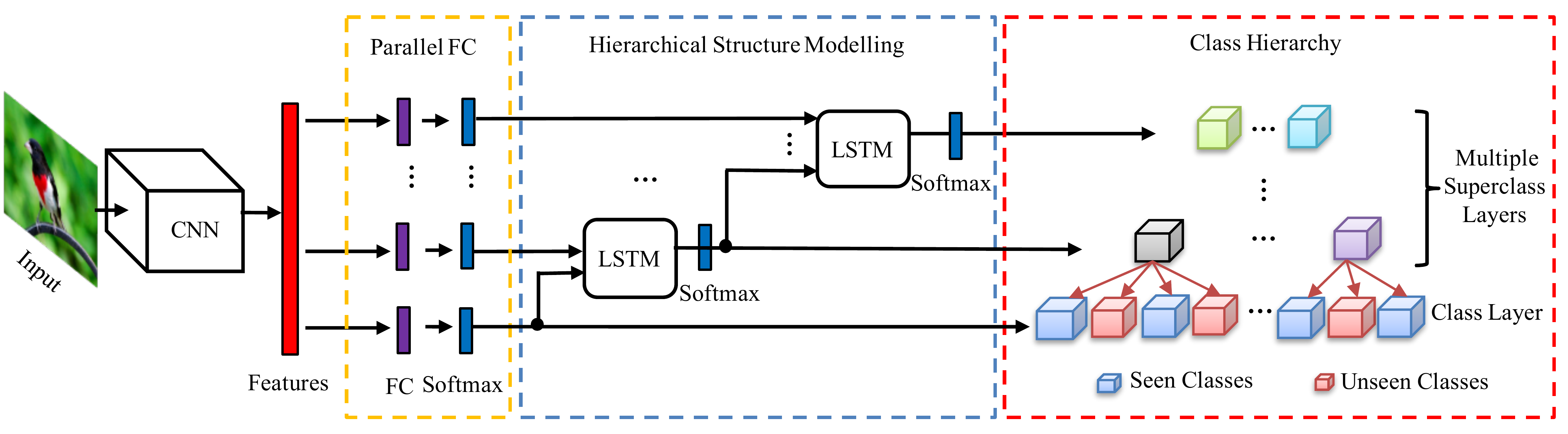}
\end{center}
\vspace{-0.1in}
\caption{Overview of the proposed CNN-RNN model for deep feature learning with three-layer class hierarchy.}
\label{feat-learn}
\vspace{-0.00in}
\end{figure*}

\section{Methodology}
\label{sect:method}

\subsection{Problem Definition}

We first formally define the ZSL problem as follows. Let $S=\{s_1,...,s_p\}$ denote the set of seen classes and $U=\{u_1,...,u_q\}$ denote the set of unseen classes, where $p$ and $q$ are the total numbers of seen classes and unseen classes, respectively. These two sets of classes are disjoint, i.e. $S\cap U=\phi$. We are given a set of labelled training images $D_s = \{(I_i, y_i, f_i,z_i):i=1,...,N_s\}$, where $I_i$ denotes the $i$-th image in the training set, $y_i\in S$ denotes its corresponding label, $f_i$ denotes its $d_f$ dimensional visual feature vector, $z_i$ denotes its $d_z$ dimensional semantic representation of the corresponding class, and $N_s$ denotes the total number of labelled images. Let $D_u = \{(I_j, y_j, f_j,z_j):j=1,...,N_u\}$ denote a set of unlabelled test images, where $I_j$ denotes the $j$-th image in the test set, $y_j\in U$ denotes its corresponding unknown labels, $f_j$ denotes its $d_f$ dimensional visual feature vector, $z_j$ denotes its $d_z$ dimensional semantic representation of the corresponding class, and $N_u$ denotes the total number of unlabelled images. We thus have the training seen class feature matrix $F_s=[f_i]_{N_s\times d_f}$, the test unseen class feature matrix $F_u=[f_j]_{N_u\times d_f}$, and the training seen class semantic matrix $Z_s=[z_i]_{N_s\times d_z}$. The goal of inductive ZSL is to predict the unseen class sample labels $\{y_j:j=1,...,N_u\}$ using a model trained with the seen class data $D_s$ only. In a generalised ZSL setting, the test samples come from both seen and unseen classes, and the ZSL problem becomes more realistic yet more challenging.

Our approach to inductive ZSL consists of three main components: class hierarchy construction, deep feature learning with superclasses, and projection function learning with superclasses. Their details are given below.

\vspace{-0.02in}
\subsection{Class Hierarchy Construction}
\label{sect:cls_hir}

In this section, we describe how to construct a tree-structured class hierarchy using a data-driven approach. The class hierarchy consists of multiple superclass layers and one single class layer. Starting from the bottom class nodes (i.e. both seen and unseen classes), we obtain the nodes in the upper layer by clustering the nodes in the lower layer, forming a tree-structured class hierarchy (see the red dashed box in Fig.~\ref{feat-learn}). We thus obtain a set $R=\{r_l: l=1,..,n_r\}$ that collects the number of clusters at each superclass layer, where $r_l$ denotes the number of clusters in the $l$-th superclass layer, and $n_r$ denotes the total number of superclass layers in the class hierarchy. Each class label $y_i$ can be mapped to its corresponding superclasses $V_i=\{v_i^l: l=1,..,n_r\}$, where $v_i^l$ denotes the superclass label of $y_i$ in the $l$-th superclass layer. This results in multiple labels at different class/superclass levels for the image ${I}_{i}$. More concretely, at each layer of the tree-structured hierarchy, we cluster the nodes by their semantic representations. Each cluster then forms a parent node (i.e. a superclass) in the upper layer of the tree. In this way, the semantic representation of the superclass is, in a sense, a mixing of the semantics of its children classes. Since the superclass labels are shared across both seen and unseen class domains, the proposed model can overcome the domain gap challenge for ZSL.

\vspace{-0.02in}
\subsection{Deep Feature Learning}
\label{sect:fea_learn}

In this section, we describe our feature extraction model, which is learned only using the seen class data $D_s$ (along with the class hierarchy constructed using semantic representations of all seen and unseen names) but expected to represent well the unseen class data $D_u$.  In this paper, we propose a CNN-RNN model defined with superclasses from the aforementioned class hierarchy to learn transferrable visual features for unseen class samples. This deep feature learning model takes an image ${I}_{i}$ as the input and then outputs a $d_f$-dimensional feature vector $f_i=\phi({I}_{i})\in \mathbb{R}^{d_f}$. Concretely, we extend a CNN model by two steps for predicting the superclass-level labels, using the shared CNN generated features $f_i$. The first step is to predict the labels at different class/superclass levels (see the yellow dashed box in Fig.~\ref{feat-learn}), so that the shared superclasses at those layers can make the learned features suitable for representing the unseen classes. The second step is to encode the hierarchical structure of class/superclass layers into superclass label prediction. That is, we infer each superclass label by considering the prediction results of the same or lower class/superclass layers (see the blue dashed box in Fig.~\ref{feat-learn}).
We are essentially learning a hierarchical classifier -- the features learned need to be useful for not only recognising the leaf-level classes, but also the higher level classes/superclasses, which are shared with the unseen classes. This makes sure that the learned feature are relevant to the unseen classes, even without using any samples from those classes. In this work, we propose to exploit the hierarchical structure using an RNN model. When RNN is combined with CNN, we obtain a CNN-RNN model for deep feature learning. We will give its technical details in the following.

For the first class/superclass label prediction step, we add $n_r+1$ parallel fully-connected (FC) layers along with softmax layers on top of the CNN model for feature extraction, as shown in the yellow dashed box in Fig.~\ref{feat-learn}. Given an object sample, each fully-connected layer with softmax thus predicts the probability distribution of the corresponding-level classes (i.e. class/superclass-level). We further introduce an RNN to model the hierarchical structure between layers and integrate this label-relation encoding module with the label prediction steps. More specifically, we model the hierarchical structure among multiple class/superclass layers by $n_r$ Long-Short-Term-Memory (LSTM) networks.

Formally, an LSTM unit has two types of states: cell state $c$ and hidden state $h$. As in \citep{lstm}, a forward pass at time $t$ with input $x_t$ is computed as follows:
\begin{eqnarray}
&&i_t=\sigma(W_{i,h}\cdot h_{t-1}+W_{i,c}\cdot c_{t-1}+W_{i,x}\cdot x_{t-1}+b_i) \nonumber\\
&&f_t=\sigma(W_{f,h}\cdot h_{t-1}+W_{f,c}\cdot c_{t-1}+W_{f,x}\cdot x_{t-1}+b_f) \nonumber\\
&&o_t=\sigma(W_{o,h}\cdot h_{t-1}+W_{o,c}\cdot c_{t-1}+W_{o,x}\cdot x_{t-1}+b_o) \nonumber \\
&&g_t=\mathrm{tanh}(W_{g,h}\cdot h_{t-1}\hspace{-0.03in}+\hspace{-0.03in}W_{g,c}\cdot c_{t-1}\hspace{-0.03in}+\hspace{-0.03in}W_{g,x}\cdot x_{t-1}\hspace{-0.03in}+\hspace{-0.03in}b_g) \nonumber \\
&&c_t=f_t*c_{t-1}+i_t*g_t \nonumber\\
&&h_t=o_t*\mathrm{tanh}(c_t)
\end{eqnarray}
where $c_t$ and $h_t$ are the models' cell and hidden states, $i_t$, $f_t$, $o_t$ are respectively the activations of input gate, forget gate, and output gate, $W_{\cdot,h}$, $W_{\cdot,c}$ are the recurrent weights, $W_{\cdot,x}$ are the input weights, and $b_{\cdot}$ are the biases. Here, $\sigma(\cdot)$ is defined with the sigmoid function.

In this paper, to model the hierarchical structure among class/superclass-level labels in the first two layers (from bottom), we use an LSTM network with two time steps. In the first time step, the output of class-level fully-connected layer (with softmax) is used as input; and in the next time step, the output of superclass-level fully-connected layer (with softmax) is used as input and the hidden cell predicts the probability distribution of superclass-level labels. The formal formulation of this LSTM network is given as:
\begin{equation}
\begin{split}
     &[h_1',c_1'] = \mathrm{LSTM}(p_0,h_0,c_0)\\
     &[h_1,c_1] = \mathrm{LSTM}(p_1,h_1',c_1')\\
     &\hat{p}_1= \mathrm{softmax}(\hat{W}_1\cdot h_1+\hat{b}_1)
\end{split}
\end{equation}
where $p_0$ denotes the output of fully-connected layer corresponding to class-level. $p_1$ denotes the output of fully-connected layer corresponding to superclass-level. $h_0$ (or $h_1'$, $h_1$) and $c_0$ (or $c_1'$, $c_1$) are the initial (or intermediate) hidden state and cell state. $\mathrm{LSTM}(\cdot)$ is a forward step of an LSTM unit. $\hat{W}_1$ and $\hat{b}_1$ are the weight and bias of the output layer, respectively. The output $\hat{p}_1$ defines a distribution over possible superclass-level labels.

Moreover, to model the hierarchical structure among classes/superclasses in the first $l~(l\geq 3)$ layers, we also employ a two-time-step LSTM network. Concretely, in the first time step, we feed the output of the previous LSTM as input, since the previous LSTM has fused all information of the first $l-1$ layers. In the next time step, the output of the current superclass-level fully-connected layer (with softmax) is used as input and the hidden cell predicts the probability distribution of the current superclass-level labels (see the blue dashed box in Fig.~\ref{feat-learn}). We thus provide the formal formulation of the LSTM network as follows:
\begin{equation}
\begin{split}
     &[h_{l-1}',c_{l-1}'] = \mathrm{LSTM}(\hat{p}_{l-1},h_{l-1},c_{l-1})\\
     &[h_l,c_l] = \mathrm{LSTM}(p_l,h_{l-1}',c_{l-1}')\\
     &\hat{p}_l = \mathrm{softmax}(\hat{W}_l\cdot h_l+\hat{b}_l)
\end{split}
\end{equation}
where $\hat{p}_{l-1}$ denotes the output of the previous LSTM. $p_l$ denotes the output of the $l$-th fully-connected layer. $h_{l-1}$ and $c_{l-1}$ are the hidden state and cell state of the previous LSTM network. $h_{l-1}'$ (or $h_{l}$) and $c_{l-1}'$ (or $c_{l}$) are the intermediate  hidden state and cell state of the current LSTM network. $\hat{W}_l$ and $\hat{b}_l$ are the corresponding weight and bias of the output layer. The output $\hat{p}_l'$ defines a distribution over possible superclass-level labels in the $l$ layer of the class hierarchy. With this method, we can obtain a total of $n_r$ two-time-step LSTM networks to model the hierarchical structure among the superclasses/classes in the hierarchy.

Therefore, by merging the hierarchical structure modelling with the original class label prediction, we define the loss function for an image $x$ as follows:
\begin{equation}
\label{eq:feat-learn}
\begin{split}
& \mathcal {L}(\theta_F,\theta_c,\theta_s)=\mathcal{L}_c(y,G_c(G(x;\theta_F);\theta_c))\\
& \hspace{0.2in} + \sum\limits_{l=1}^{n_r}\lambda_l \mathcal{L}_{s_l}(v_l,G_{s_l}(G(x;\theta_F);\theta_{s_l}))
\end{split}
\end{equation}
where $\mathcal {L}_{s_l}$($G_{s_l}$, $\theta_{s_l}$) denotes the loss (network, parameters) of the $l$-th LSTM which models the hierarchical structure among classes/superclasses in the first $l+1$ layer. $y$ and $\{v_l:l=1,...,n_r\}$ denote the true label of the image at class level and $n_r$ superclass levels, respectively. $\theta_F$ (or $G$) denotes the parameters (or network) of the CNN. $\mathcal {L}_c$ ($G_c$, $\theta_c$) denotes the loss (network, parameters) of class-level label prediction. $\lambda_l$ weights these losses.

\subsection{Projection Function Learning}
\label{sect:zsl_method}

Once the feature extraction model is learned using the training seen class data, it can be used to extract visual features from both training and test images and then compute the two feature matrices $F_s$ and $F_u$ for the seen and unseen class domains, respectively. With the seen class feature matrix $F_s$, we propose a novel projection learning method which exploits the superclasses from the class hierarchy to align the seen and unseen class domains. Since the superclasses are shared across the two domains, the proposed method enables us to learn a transferrable projection function for inductive ZSL. We give the details of our method below.

\subsubsection{Projection Learning over Superclasses}

As mentioned in Sec.~\ref{sect:cls_hir}, each class label $y_i$ can be mapped to its corresponding superclasses $V_i=\{v_i^l: l=1,..,n_r\}$, where $v_i^l$ denotes the superclass label of $y_i$ in the $l$-th superclass layer of the class hierarchy. Let $e^l_i$ denote the $d_z$ dimensional semantic vector of the superclass label $v_i^l$. We thus obtain $n_r$ training superclass semantic matrices $\{E_s^l=[e^l_i]_{N_s\times d_z},l=1,...,n_r\}$, where each matrix collects the semantic vectors of superclasses in the corresponding layer for all seen class samples.

Given the training seen class feature matrix $F_s \in \mathbb{R}^{N_s\times d_f}$ and the initial  training superclass semantic matrix $E_s^l\in \mathbb{R}^{N_s\times d_z}$, we model the set of seen class images as a graph $\mathcal{G}=\{\mathcal{W},V\}$ with its vertex set $V=F_s$ and weight matrix $\mathcal{W}=\{w_{uv}\}$, where $w_{uv}$ denotes the similarity between visual features of the $u$-th and $v$-th seen class images (i.e. $f_u$ and $f_v$). It should be noted that the weight matrix $\mathcal{W}$ is usually assumed to be nonnegative and symmetric. We thus compute the normalised Laplacian matrix $\mathcal{L}$ of the graph $\mathcal{G}$ by the following equation:
\begin{equation}
\mathcal{L}=I-D^{-1/2}\mathcal{W}D^{-1/2}
\label{Eq:graph}
\end{equation}
where $I$ is an $N_s\times N_s$ identity matrix, and $D$ is an $N_s\times N_s$
diagonal matrix with its $u$-th diagonal entry $\sum_vw_{uv}$.

Based on the well-known Laplacian regularisation and superclass semantic representations, our projection learning method solves the following optimisation problem with respect to the $l$-th superclass layer:
\begin{equation}
\begin{split}
& \min\limits_{W^l,\widetilde{E}_s^l}L(W^l,\widetilde{E}_s^l)\\
&=\|F_sW^l-\widetilde{E}_s^l\|_F^2+\mu^l \|F_s-\widetilde{E}_s^lW^{lT}\|_F^2\\
&+\epsilon^l tr(\widetilde{E}_s^{lT}\mathcal{L}\widetilde{E}_s^l)+\nu^l\|\widetilde{E}_s^l-E_s^l\|_F^2 +\eta^l\|W^l\|_F^2
\label{Eq:projection}
\end{split}
\end{equation}
where $W^l \in \mathbb{R}^{d_f\times d_z}$ is the projection matrix between the superclass semantic representation corresponding to the $l$-th superclass layer and the visual feature representation, and $\widetilde{E}_s^l\in \mathbb{R}^{N_s\times d_z}$ is the updated training superclass semantic matrix during model learning.

In the above objective function, the first two terms aim to learn bidirectional linear projections between $F_s$ and $\widetilde{E}_s^l$, which in fact is a self-reconstruction task \citep{Kodirov2017CVPR} and has been verified to have good generalisation ability. Moreover, the third term denotes the well-known graph regularisation, which can enforce the superclass semantic representations $\widetilde{E}_s^l$ to preserve the graph locality of image features $F_s$, which benefits the feature self-reconstruction. The fourth term is a fitting constraint between $\widetilde{E}_s^l$ and $E_s^l$, meaning the the instance-level semantic representations should be close to their superclass prototype. The last term is the Frobenius norm used to regularise $W^l$. These terms are weighed by the four regularisation parameters $\mu^l,\epsilon^l,\nu^l,\eta^l$.

We further develop an efficient approach to tackle the graph regularised self-reconstruction problem in Eq.~(\ref{Eq:projection}). Specifically, we solve Eq.~(\ref{Eq:projection}) by alternately optimising the following two subproblems:
\begin{equation}
W^{l*}=\arg\min\limits_{W^l} L(W^l,\widetilde{E}_s^{l*})
\label{Eq:opt1}
\end{equation}
\begin{equation}
\widetilde{E}_s^{l*}=\arg\min\limits_{\widetilde{E}^l_s} L(W^{l*},\widetilde{E}^l_s)
\label{Eq:opt2}
\end{equation}
Here, $\widetilde{E}_s^{l*}$ is initialised with $E_s^l$. Taking a convex quadratic formulation, each of the above two subproblems has a global optimal solution. The two solvers are given below.

For the first subproblem, with $\widetilde{E}^l_s=\widetilde{E}_s^{l*}$ fixed, the solution of $\arg\min\limits_{W^l} L(W^l,\widetilde{E}_s^{l*})$ can be found by setting $\frac{\partial L(W^l, \widetilde{E}_s^{l*})}{\partial W^l}=0$. We thus obtain a linear equation:
\begin{equation}
(F_s^TF_s+\eta^l I)W^l+\mu^l W^l(E_s^{l*})^TE_s^{l*}=(1+\mu^l) F_s^TE_s^{l*}
\end{equation}
Let $\alpha^l = \mu^l/(1+\mu^l) \in (0, 1)$ and $\gamma^l = \eta^l/(1+\mu^l)$. In this paper, we empirically set $\gamma^l=0.01$. We have:
\begin{equation}
[(1-\alpha^l)F_s^TF_s+\gamma^l I]W^l + W^l(\alpha^l (E_s^{l*})^TE_s^{l*})= F_s^TE_s^{l*}
\label{eq:bpl}
\end{equation}
which can be viewed as a Sylvester equation. Since $(1-\alpha^l)F_s^TF_s+\gamma^l I \in\mathbb{R}^{d_f\times d_f}$ and $\alpha^l (E_s^{l*})^TE_s^{l*}\in \mathbb{R}^{d_z\times d_z}$ ($d_f,d_z\ll N_s$), this equation can be solved efficiently by the Bartels-Stewart algorithm \citep{BS-alg}.

\begin{algorithm}[t]
\caption{Projection Learning over Superclasses}
\label{alg:pfl}
\begin{algorithmic}\smallskip
\STATE {\bfseries Input:} ~Training seen class feature matrix $F_s$\\
\qquad~~~~Initial training  superclass semantic matrix $E_s^l$\\
\qquad~~~~Parameters $\alpha^l,\beta^l, \epsilon^l$
\STATE {\bfseries Output:} $W^{l*}$
\STATE 1. Set $\widetilde{E}_s^{l*} = E_s^l$;
\STATE 2. Construct a graph $\mathcal{G}$ and compute $\mathcal{L}$ with Eq.~(\ref{Eq:graph});
\REPEAT
\STATE 3. Find the best solution $W^{l*}$ by solving Eq.~(\ref{eq:bpl});
\STATE 4. Updating $\widetilde{E}^l_s$ with $W^{l*}$ by solving Eq.~(\ref{eq:ssl});
\UNTIL{a stopping criterion is met}
\smallskip
\end{algorithmic}
\end{algorithm}

For the second subproblem, with $W^l = W^{l*}$ fixed, the solution of $\arg \min_{\widetilde{E}_s^l} L(W^{l*},\widetilde{E}_s^l)$ can be found by setting $\frac{\partial L(W^{l*},\widetilde{E}_s^l)}{\partial \widetilde{E}_s^l}=0$. We thus have:
\begin{equation}
\begin{split}
&\widetilde{E}_s^l(\mu^l (W^{l*})^TW^{l*}+(1+\nu^l)I)+\epsilon^l \mathcal {L}\widetilde{E}_s^l\\
&\hspace{0.0in}=(1+\mu^l)F_sW^{l*}+\nu^l E_s^l
\end{split}
\end{equation}
Let $\beta^l = 1/(1+\nu^l) \in (0, 1)$. We obtain:
\begin{equation}\begin{split}
&\widetilde{E}_s^l[\alpha^l\beta^l (W^{l*})^TW^{l*}+(1-\alpha^l)I]+\epsilon^l \mathcal {L}\widetilde{E}_s^l\\
&\hspace{0.0in}=\beta^l F_sW^{l*}+(1-\alpha^l)(1-\beta^l) E_s^l
\end{split}
\label{eq:ssl}
\end{equation}
which is a Sylvester equation. Since $\alpha^l\beta^l (W^{l*})^TW^{l*}+ (1-\alpha^l)I \in \mathbb{R}^{d_z\times d_z}$ ($d_z \ll N_s$), this equation can be solved efficiently by the Bartels-Stewart algorithm. Importantly, the time complexity of solving Eqs.~(\ref{eq:bpl}) and (\ref{eq:ssl}) is linear with respect to the number of samples. Given that the proposed algorithm is shown to converge very quickly, it is efficient even for large-scale problems.

To sum up, we outline the proposed projection function learning method in Algorithm \ref{alg:pfl}. Once the best projection function is learned, we first project the superclass prototypes corresponding to the $l$-th superclass layer into the feature space. We then perform nearest neighbour search over the possible superclasses (obtained from the higher-level superclass layer as described in Sec.~\ref{sec:lab_infer}) in the feature space to predict the superclass labels corresponding to the $l$-th superclass layer for test samples.

Note that we do not directly exploit the CNN-RNN model proposed in Sec.~\ref{sect:fea_learn} to predict the superclass labels of the test unseen class samples, because \emph{this model would fail when some superclasses only contain unseen classes}\footnote{Our CNN-RNN model can still extract transferrable features for these test unseen class samples because the corresponding unseen classes share higher-level superclasses with some seen classes.}. In contrast, our projection learning method addresses this issue by projecting the superclass semantic representations to the feature space and then performing nearest neighbor search over these superclass prototypes for label inference.

\vspace{-0.1in}
\subsubsection{Full ZSL Algorithm}
\label{sec:lab_infer}

The results of the proposed projection learning method can be used for inferring the labels of unseen class samples as follows. First, with respect to the current layer of the class hierarchy, we predict the top 3 superclass labels of each test unlabelled unseen sample $x_j$ using the optimal projection matrix learned by Algorithm \ref{alg:pfl}. Second, derived from the top 3 superclass labels of $x_j$, we obtain the set of the most possible superclass labels of $x_j$ according to the lower-level superclass layer of the class hierarchy. In the same way, the set of the most possible unseen class labels $\mathcal{N}(x_j)$ can be acquired for the single class/leaf layer. Finally, we learn the projection function using seen class samples to infer the labels of unseen class samples by solving:
\begin{equation}
\begin{split}
& \min\limits_{W,\widetilde{Z}_s}L(W,\widetilde{Z}_s)\\
&=\|F_sW-\widetilde{Z}_s\|_F^2+\mu^c \|F_s-\widetilde{Z}_sW^T\|_F^2\\
&+\epsilon^c tr(\widetilde{Z}_s^T\mathcal{L}\widetilde{Z}_s)+\nu^c\|\widetilde{Z}_s-Z_s\|_F^2 +\eta^c\|W\|_F^2
\label{Eq:projection_cls}
\end{split}
\end{equation}
where $W \in \mathbb{R}^{d_f\times d_z}$ is the projection matrix, and $\widetilde{Z}_s\in \mathbb{R}^{N_s\times d_z}$ is the updated training seen class semantic matrix during model learning. To solve this optimisation problem, we can develop a solver similar to Algorithm \ref{alg:pfl} (also need to tune parameters $\alpha^c$, $\beta^c$ and $\epsilon^c$ as Algorithm \ref{alg:pfl}) and the only difference is that the training superclass semantic matrix is replaced by the class-level semantic matrix. When the best projection matrix is learned, we can project the class-level semantic prototypes from the test set into the feature space. The nearest neighbor search is then performed (over the set of the most possible unseen class labels) in the feature space to predict the label of a test image.

\begin{algorithm}[t]
\caption{Full ZSL Algorithm}
\label{alg:fzsl}
\begin{algorithmic}
\smallskip
\STATE {\bfseries Input:} Training set of seen class images $D_s$\\
\qquad~~~~Test set of unseen class images $D_u$\\
\qquad~~~~Parameters $\lambda_l$, $\alpha^l,\beta^l,\epsilon^l$,$\alpha^c,\beta^c,\epsilon^c$
\STATE {\bfseries Output:} Labels of test samples
\STATE {\bfseries Class Hierarchy Construction:}
\STATE 1. Construct the tree-structured class hierarchy as in Sec.~\ref{sect:cls_hir};
\STATE {\bfseries Deep Feature Learning:}
\STATE 2. Train a CNN-RNN model according to Eq. (\ref{eq:feat-learn});
\STATE 3. Compute feature matrices $F_s$ for the seen class domain;
\STATE {\bfseries Projection Function Learning:}
\STATE 4. Solve Eq.~(\ref{Eq:projection}) for predicting the superclass labels using Algorithm~\ref{alg:pfl};
\STATE 5. Generate the set of the most possible unseen class labels for \\each test sample;
\STATE 6. Solve Eq.~(\ref{Eq:projection_cls}) with an algorithm similar to Algorithm \ref{alg:pfl};
\STATE 7. Predict the unseen class label of each test sample.
\smallskip
\end{algorithmic}
\end{algorithm}

Different from existing projection learning methods that mainly rely on class-level semantics for label inference, we exploit both class-level and superclass-level semantics to recognise unseen class samples. Since the superclasses in our hierarchy are shared across the two domains, our projection learning method can alleviate the projection domain shift and thus benefit unseen class image recognition. By combining class hierarchy construction, deep feature learning, and projection function learning together for inductive ZSL, our full algorithm is outlined in Algorithm \ref{alg:fzsl}.

\vspace{-0.1in}
\subsection{Extension to Few-Shot Learning}

Although the proposed model is originally designed for ZSL, it can be easily extended to FSL. Under a standard FSL setting, the dataset is split into three parts: a training set of many labelled base/seen class samples, a support set of few labelled novel/unseen class samples, and a test set of the rest novel class samples. In this work, we first construct a tree-structured class hierarchy using all base and novel class prototypes as in Sec.~\ref{sect:cls_hir}. With the obtained class hierarchy, we further train our full model (including deep feature learning and projection function learning) over the whole training set as before, and predict the labels of test samples as in Sec.~\ref{sec:lab_infer}. To obtain better FSL results, we exploit both average visual features of few shot samples per novel class and the projected novel class prototypes for nearest neighbor search in the feature space.

\begin{table}[t]
\vspace{0.00in}
\caption{Details of four benchmark datasets. Notations: `SS' -- semantic space, `SS-D' -- the dimension of semantic space, `A' -- attribute, and `W' -- word vector. }
\label{dataset}
\vspace{-0.00in}
\begin{center}
\tabcolsep0.3cm
\begin{tabular}{l cccc}
\hline\noalign{\smallskip}
Dataset		 & \#instances &SS&SS-D &\#seen/unseen\\\noalign{\smallskip}
\hline\noalign{\smallskip}
AwA		&30,475&A&85&40/10\\
CUB		&11,788&A&312&150/50\\
SUN		&14,340&A&102&645/72\\
ImNet	&218,000&W&1,000&1,000/360\\
\noalign{\smallskip}\hline
\end{tabular}
\end{center}
\vspace{-0.05in}
\end{table}

\vspace{-0.08in}
\section{Experiments}

\vspace{-0.05in}
\subsection{Zero-Shot Learning}

\vspace{-0.05in}
\subsubsection{Datasets and Settings}
\label{zsl_exp}

\noindent\textbf{Datasets}. Four widely-used benchmark datasets are selected. Three of them are of medium-size: Animals with Attributes (AwA) \citep{Lampert2014pami}, Caltech UCSD Birds (CUB) \citep{CUB-200-2011}, and SUN Attribute (SUN) \citep{Patterson2014ijcv}. One large-scale dataset is ILSVRC2012/2010 (ImNet)\citep{Russakovsky2015ImageNet}, where the 1,000 classes of ILSVRC2012 are used as seen classes and 360 classes of ILSVRC2010 (not included in ILSVRC2012) are used as unseen classes, as in \citep{Fu2016CVPR}. The details of these benchmark datasets are given in Table \ref{dataset}.

\noindent\textbf{Semantic Space}. We use two types of semantic spaces. For the three medium-scale datasets, attributes are used as the semantic representations. For the large-scale ImNet dataset, the semantic word vectors are employed to form the semantic space. In this paper, we train a skip-gram text model on a corpus of 4.6M Wikipedia documents to obtain the word2vec \citep{Norouzi14iclr} word vectors.

\noindent\textbf{ZSL Settings}. (1) Standard ZSL: This ZSL setting is used in pre-2017 works \citep{Kodirov2015ICCV,Kodirov2017CVPR,Zhang2015iccv,zhang2016cvpr,Akata2015CVPR,zhang2016eccv}. Concretely, the train/test (or seen/unseen classes) splits of 40/10, 150/50, 645/72, and 1,000/360 are provided for AwA, CUB, SUN, and ImNet, respectively. (2) Pure ZSL: A new `pure' ZSL setting \citep{Xian2017CVPR} was proposed to overcome the weakness in the standard ZSL setting. Specifically, most recent ZSL models extract the visual features using ImageNet ILSVRC2012 1K class pretrained CNN models, but the unseen classes of the three medium-scale datasets in the standard splits may overlap with the 1K ImageNet classes. The zero-shot rule is thus violated. Under the new ZSL setting, new benchmark splits are provided to ensure that the unseen classes have no overlap with the ImageNet ILSVRC2012 1K classes. (3) Generalised ZSL: Another recently appearing setting is the generalised ZSL setting \citep{Xian2017CVPR}, under which the test set contains data samples from both seen and unseen classes. This setting is more suitable for real-world applications.

\noindent\textbf{Evaluation Metrics}. (1) Standard ZSL: For three medium-scale datasets, we compute the top-1 classification accuracy over all test samples as in previous works \citep{Kodirov2015ICCV,Zhang2015iccv,zhang2016cvpr}. (2) Pure ZSL: For three medium-scale datasets, we compute average per-class top-1 accuracy as in \citep{Xian2017CVPR}. For the large-scale ImNet dataset, the flat h@5 classification accuracy is computed as in \citep{Fu2016CVPR,Kodirov2017CVPR}, where h@5 means that a test image is classified to a 'correct label' if it is among the top five labels. (3) Generalised ZSL: Three evaluation metrics are defined: 1) $acc_s$ -- the accuracy of classifying the data samples from the seen classes to all the classes (both seen and unseen); 2) $acc_u$ -- the accuracy of classifying the data samples from the unseen classes to all the classes; 3) HM -- the harmonic mean of $acc_s$ and $acc_u$.

\noindent\textbf{Compared Methods}. A wide range of existing ZSL models are selected for comparison. Under each ZSL setting, we focus on the recent and representative ZSL models that have achieved the state-of-the-art results.

\subsubsection{Implementation Details}

\begin{table*}[ht]
\vspace{0.00in}
\centering
\caption{Comparative accuracies (\%) of standard ZSL. Visual features: GOO -- GoogLeNet \citep{szegedy2015cvpr}; VGG -- VGG Net \citep{simonyan2014arxiv}; RES -- ResNet \citep{he2016cvpr}; Feat-Learn -- deep feature learning. }
\label{res_st}
\vspace{-0.04in}
\tabcolsep0.42cm
\begin{small}
\begin{tabular}{l c c c c c}
\hline\noalign{\smallskip}
Model&Visual Features &Inductive?&AwA&CUB&SUN\\
\noalign{\smallskip}\hline\noalign{\smallskip}
SP-ZSR \citep{zhang2016eccv} & VGG   &  no & 92.1 & 55.3 &  -- \\
SSZSL \citep{shojaee2016semi}	&VGG		 &no&88.6&58.8&-- \\
DSRL \citep{ye2017cvpr}   	&VGG		 &no&87.2&57.1&--   \\
TSTD \citep{yu2017transductive}&VGG	 &no&90.3&58.2&--    \\
BiDiLEL \citep{wang2017ijcv}  &VGG	 &no&\textbf{95.0}&62.8  &--   \\
DMaP \citep{Li2017CVPR} & VGG+GOO+RES   & no & 90.5 & \textbf{67.7} &  --\\
\noalign{\smallskip}\hline\hline\noalign{\smallskip}
JLSE \citep{zhang2016cvpr}	&VGG		 	&yes	&80.5&42.1 &--   \\
LAD \citep{jiang2017iccv}    	&VGG	 	&yes	&82.5&56.6&--   \\
SCoRe \citep{Morgado2017cvpr}	&VGG	 	&yes	&82.8&59.5&--    \\
LESD \citep{Ding2017cvpr}		&VGG	 	&yes	&82.8&56.2&--   \\
SJE \citep{Akata2015CVPR}		&GOO &yes	&73.9&51.7 &56.1\\
SynC \citep{Changpinyo2016CVPR}&GOO 	&yes	&72.9&54.7 &62.7\\
VZSL \citep{Wang2018aaai}&VGG &yes&85.3&57.4&--\\
SAE \citep{Kodirov2017CVPR} 	&GOO 	&yes	&84.7&61.4&65.2\\
CVAE \citep{Mishra2017} & RES   & yes & 85.8 & 54.3  &--\\
CLN+KRR \citep{long2017zero} &Feat-Learn &yes	& 81.0&58.6&--   \\
GVR \citep{bucher2017iccv} 	&Feat-Learn &yes	&87.8&60.1&56.4\\
\noalign{\smallskip}\hline\noalign{\smallskip}
Ours  &Feat-Learn 	&yes	 &\textbf{91.7} & \textbf{67.2} & \textbf{69.6}\\
\noalign{\smallskip}\hline
\end{tabular}
\end{small}
\end{table*}

\begin{table*}[t]
%\vspace{0.2in}
\caption{Comparative accuracies (\%) of pure ZSL. For ImNet, the hit@5 accuracy is used. The notations are the same as in Table~\ref{res_st}. }
\label{res_pu}
\vspace{-0.04in}
\begin{center}
\tabcolsep0.6cm
\begin{tabular}{lccccc}
\hline\noalign{\smallskip}
Model		&Visual Features&AwA&CUB&SUN & ImNet\\
\noalign{\smallskip}\hline\noalign{\smallskip}
CMT \citep{socher2013nips}	&RES&39.5&34.6&39.9&--\\
DeViSE \citep{Frome2013nips} &RES&54.2&52.0&56.5&12.8\\
DAP \citep{Lampert2014pami}	&RES&44.1&40.0&39.9&--\\
ConSE \citep{Norouzi14iclr}	&RES&45.6&34.3&38.8&15.5\\
SSE \citep{Zhang2015iccv} 	&RES&60.1&43.9&51.5&--\\
SJE \citep{Akata2015CVPR}		&RES&65.6&53.9&53.7&--\\
ALE \citep{zhang2016cvpr}		&RES&59.9&54.9&58.1&--\\
SynC \citep{Changpinyo2016CVPR}&RES&54.0&55.6&56.3&--\\
SP-AEN \citep{chen2018cvpr}&RES&58.5&55.4&59.2&--\\
CVAE \citep{Mishra2017}&VGG&71.4&52.1&61.7&24.7\\
SAE \citep{Kodirov2017CVPR}&GOO&61.3&48.2&59.2&27.2\\
DEM \citep{Zhang2017cvpr}&GOO&68.4&51.7&61.9 & 25.7\\
CLN+KRR \citep{long2017zero} &Feat-Learn&68.2&58.1&60.0&--\\
WGAN+ALE \citep{Xian2018cvpr} &Feat-Learn&68.2&61.5&62.0&--\\
\noalign{\smallskip}\hline\noalign{\smallskip}
Ours &Feat-Learn&\textbf{72.2}&\textbf{65.8}&\textbf{62.9}&\textbf{28.6}\\
\noalign{\smallskip}\hline
\end{tabular}
\end{center}
\vspace{-0.02in}
\end{table*}

\begin{figure}[t]
\vspace{0.00in}
\includegraphics[width=0.88\columnwidth]{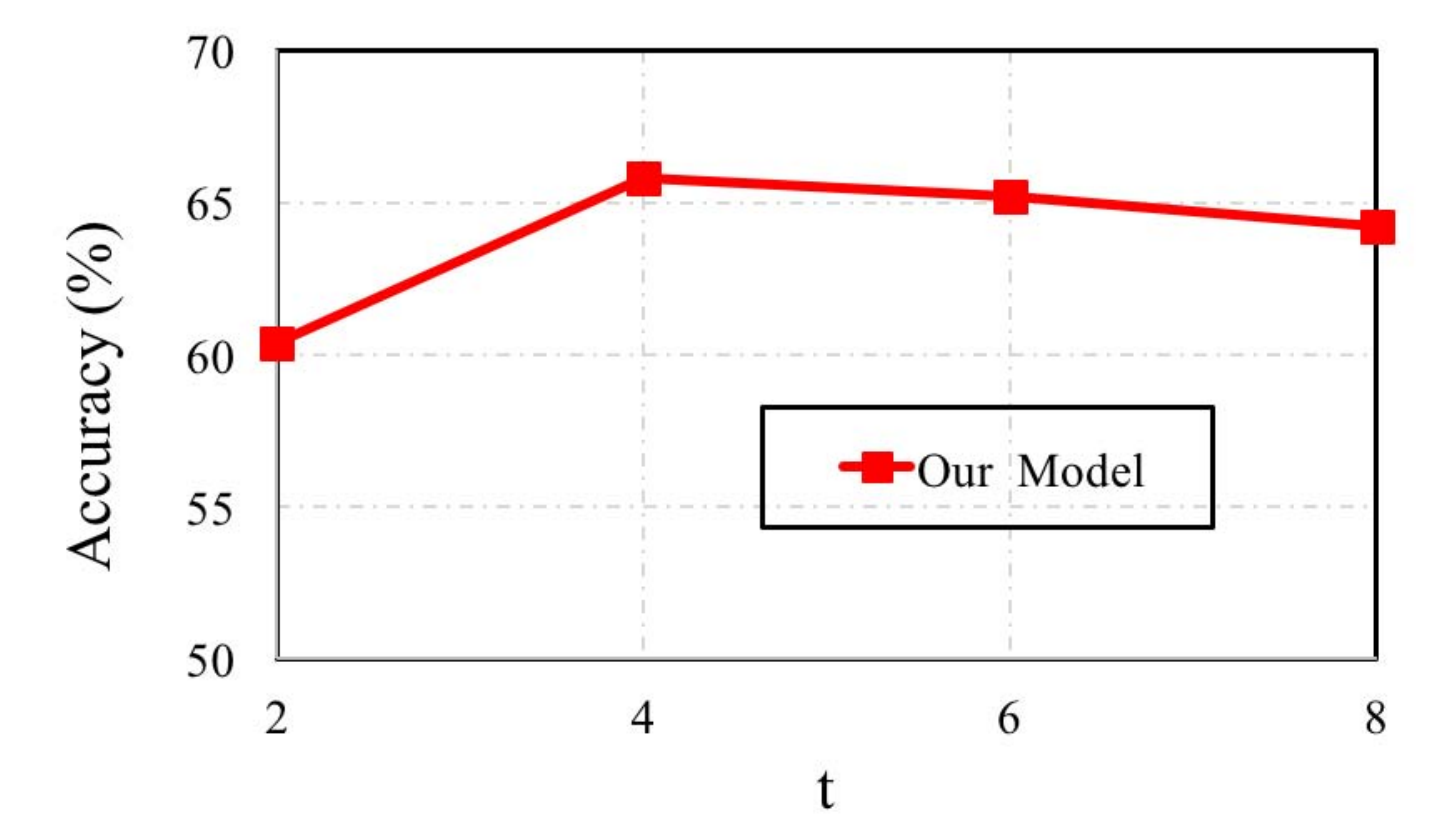}
\vspace{-0.05in}
\caption{Illustration of the effect of the parameter $t$ on our model under the pure ZSL setting for the CUB dataset.}
\vspace{-0.08in}
\label{res_t}
\end{figure}

\noindent\textbf{Class Hierarchy Construction}. $R=\{r_l: l=1,..,n_r\}$ that collects the number of clusters (i.e. $r_l$) at each superclass layer of the tree-structured class hierarchy (with $n_r$ superclass layers) is empirically defined as: $r_l = \lfloor (p+q)/t^{l-2}\rfloor$. $n_r$ is determined with the constraint that each layer has at least $t$ nodes. To study the influence of $t$ on our model, we provide the results of our model with different hierarchy structures (determined by the parameter $t$) in Fig.~\ref{res_t}. We can observe that the performance will drop when $t$ becomes oversize (e.g. $t=8$). It is reasonable that the oversize parameter $t$ leads to a very shallow hierarchy and the improvements brought by our model thus diminish. Therefore, to better select the parameter $t$, we perform cross-validation on the training data, as in  \citep{Kodirov2017CVPR}.

\noindent\textbf{Deep Network Training}. In our CNN-RNN model, the first five groups of convolutional layers in VGG-16 Net \citep{simonyan2014arxiv} are used as the CNN subnet. The convolutional layers of this CNN subnet are pre-trained on ILSVRC 2012 \citep{Russakovsky2015ImageNet}, while the other layers (including the LSTMs) are trained from scratch. Stochastic gradient descent (SGD) \citep{sgd} is used for model training with a base learning rate of 0.001. For those layers trained from scratch, their learning rates are 10 times of the base learning rate. The deep learning is implemented with Caffe \citep{jia2014caffe}.

\noindent\textbf{Projection Function Learning}. Our projection function learning model has three groups of parameters to tune: $\{\alpha^l:l=1,...,n_r\}$ (in Eq.~(\ref{eq:bpl})), $\{\beta^l:l=1,...,n_r\}$ (in Eq.~(\ref{eq:ssl})), and $\{\epsilon^l:l=1,...,n_r\}$ (in Eq.~(\ref{eq:ssl})). In this paper, we select $\alpha^1,\beta^1,\epsilon^1$ by cross-validation on the training data as in \citep{Kodirov2017CVPR}. Then, we directly use them in other superclass-level projection learning and class-level projection learning (in Eq.~(\ref{Eq:projection_cls})).

\subsubsection{Comparative Evaluation}

\begin{figure}[t]
\vspace{0.05in}
\begin{center}
\includegraphics[width=0.88\columnwidth]{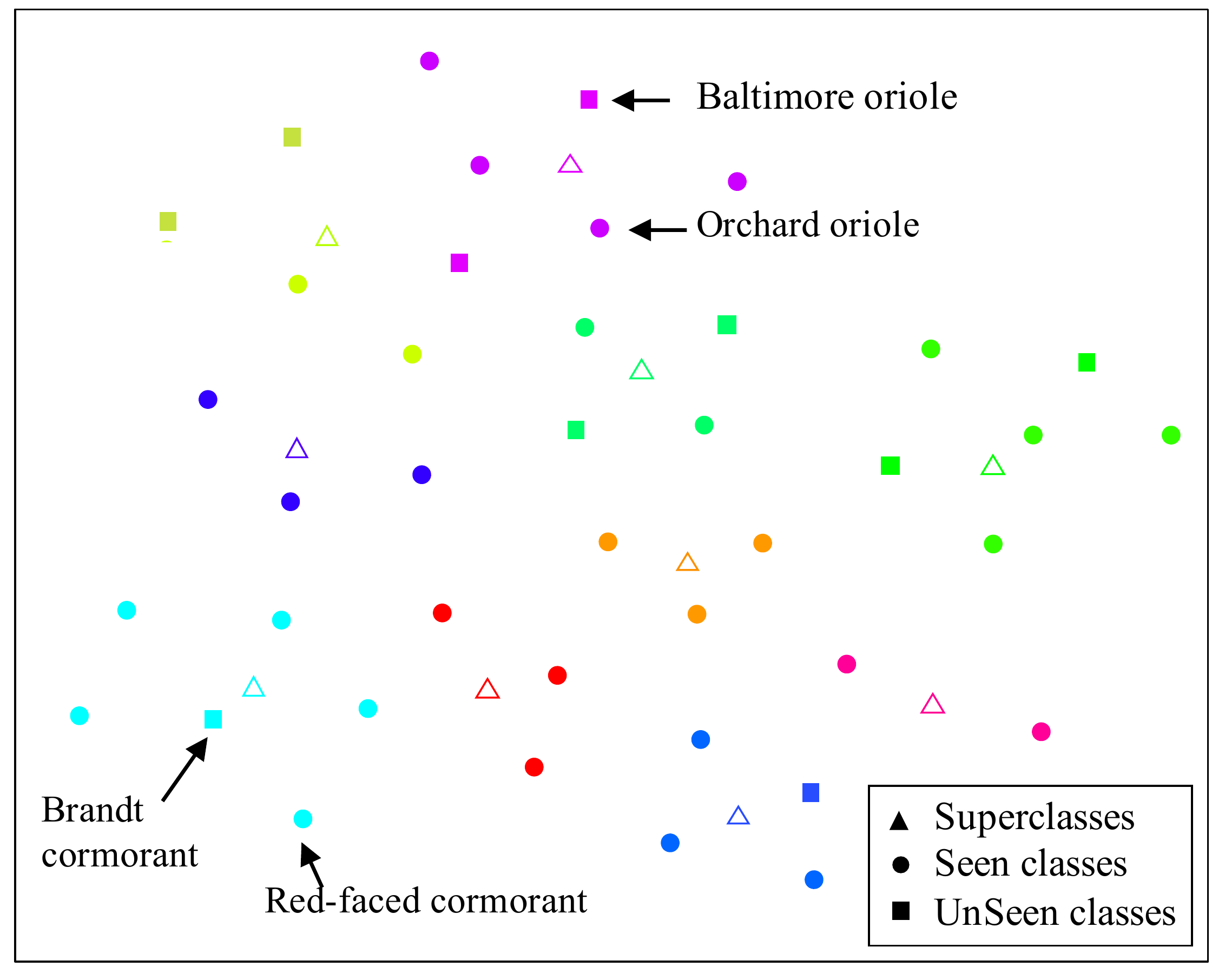}
\end{center}
\vspace{-0.00in}
\caption{The t-SNE visualisation of the superclasses generated by clustering on the CUB dataset. For easy viewing, we randomly choose 10 superclasses (marked with different colors).}
\label{cluster}
\vspace{0.1in}
\end{figure}

\begin{table*}[t]
\vspace{0.05in}
\caption{Comparative results (\%) of generalised ZSL. The \emph{overall} performance is evaluated by the HM metric.}
\label{res_gz}
\vspace{-0.0in}
\begin{center}
\tabcolsep0.3cm
\begin{tabular}{lccccccccc}
\hline\noalign{\smallskip}
\multirow{2}{*}{Model}&\multicolumn{3}{c}{AwA}&\multicolumn{3}{c}{CUB}&\multicolumn{3}{c}{SUN}\\\noalign{\smallskip}\cline{2-10}\noalign{\smallskip}
&$acc_s$&$acc_u$&HM&$acc_s$&$acc_u$&HM&$acc_s$&$acc_u$&HM\\
\noalign{\smallskip}\hline\noalign{\smallskip}
CMT \citep{socher2013nips}	&86.9&8.4&15.3&60.1&4.7&8.7&28.0&8.7&13.3\\
DeViSE \citep{Frome2013nips}&68.7&13.4&22.4&53.0 &23.8&32.8&27.4&16.9&20.9\\
SSE \citep{Zhang2015iccv}&80.5&7.0&12.9&46.9&8.5&14.4&36.4&2.1&4.0\\
SJE \citep{Akata2015CVPR}  	&74.6&11.3&19.6&59.2&23.5&33.6&30.5&14.7&19.8\\
LATEM \citep{Xian16}  	&71.7&7.3&13.3&57.3&15.2&24.0&28.8&14.7&19.5\\
ALE \citep{akata2016pami}&76.1&16.8&27.5&62.8&23.7&34.4&33.1&21.8&26.3\\
ESZSL\citep{Romera2015icml}&75.6&6.6&12.1&63.8&12.6&21.0&27.9&11.0&15.8\\
SynC\citep{Changpinyo2016CVPR}&87.3&8.9&16.2&\textbf{70.9}&11.5&19.8&\textbf{43.3}&7.9&13.4\\
SAE\citep{Kodirov2017CVPR}&71.3&31.5&43.5&36.1&28.0&31.5&25.0&15.8&19.4\\
DEM\citep{Zhang2017cvpr}&84.7&32.8&47.3&57.9&19.6&29.2&34.3&20.5&25.6\\
SP-AEN\citep{chen2018cvpr}&\textbf{90.9}&23.3&37.1&70.6&34.7&46.6&38.6&24.9&30.3\\
WGAN+ALE \citep{Xian2018cvpr} &57.2&{\bf 47.6}&52.0&59.3&40.2&47.9&31.1&{\bf41.3}&{\bf35.5}\\
\noalign{\smallskip}\hline\noalign{\smallskip}
Ours&68.2&44.5&{\bf53.9}&46.9&{\bf51.1}&{\bf48.9}&30.2&31.4&30.7\\
\noalign{\smallskip}\hline
\end{tabular}
\end{center}
\end{table*}

\noindent\textbf{Standard ZSL}. Table~\ref{res_st} shows the results of the comparison to the state-of-the-art ZSL models under the standard ZSL setting. We can obtain the following observations: (1) Our full model for inductive ZSL clearly outperforms the state-of-the-art inductive alternatives. Particularly, the improvements obtained by our model over the strongest inductive competitor are significant for AwA, CUB, and SUN, i.e. 3.9\%, 5.8\%, and 4.4\%, respectively. This validates the effectiveness of our model for inductive ZSL. (2) Our model even yields significantly better results than some recent transductive ZSL models. This is really impressive given that these transductive models have access to the full test data for model training and thus have an inherent advantage over the inductive ZSL models including ours. (3) All recent inductive ZSL models that induce deep feature learning into ZSL are quite competitive, as compared to those without deep feature learning. Among these deep feature learning models, our model clearly performs the best thanks to transferrable feature learning with the superclasses from the proposed class hierarchy.

\noindent\textbf{Pure ZSL}. By following the same `pure' ZSL setting in \citep{Xian2017CVPR}, the overlapped ImageNet ILSVRC2012 1K classes are removed from the test set of unseen classes for the three medium-scale datasets, while the split of the ImNet dataset naturally satisfies the `pure' ZSL setting. Table \ref{res_pu} presents the comparative results under the pure ZSL setting. It can be seen that: (1) Our model yields the best results on all four datasets and achieves about 1-4\% performance improvement over the best competitors. This validates the effectiveness of our model for this stricter ZSL setting. (2) On the large-scale ImNet dataset, our model achieves about 1--4\% improvements over the state-of-the-art ZSL models \citep{Mishra2017,Kodirov2017CVPR,Zhang2017cvpr}, showing the scalability of our model for large-scale ZSL problems. (3) Our model clearly outperforms the state-of-the-art feature learning model \citep{Xian2018cvpr}, due to our transferrable feature and projection learning with the superclasses. This is also supported by the t-SNE visualisation of the superclasses in Fig.~\ref{cluster}, where an unseen class tends to be semantically related to a seen class within the same superclass (i.e. the superclasses are shared across the seen and unseen domains).

\begin{figure}[t]
\vspace{0.1in}
\begin{center}
\includegraphics[width=0.98\columnwidth]{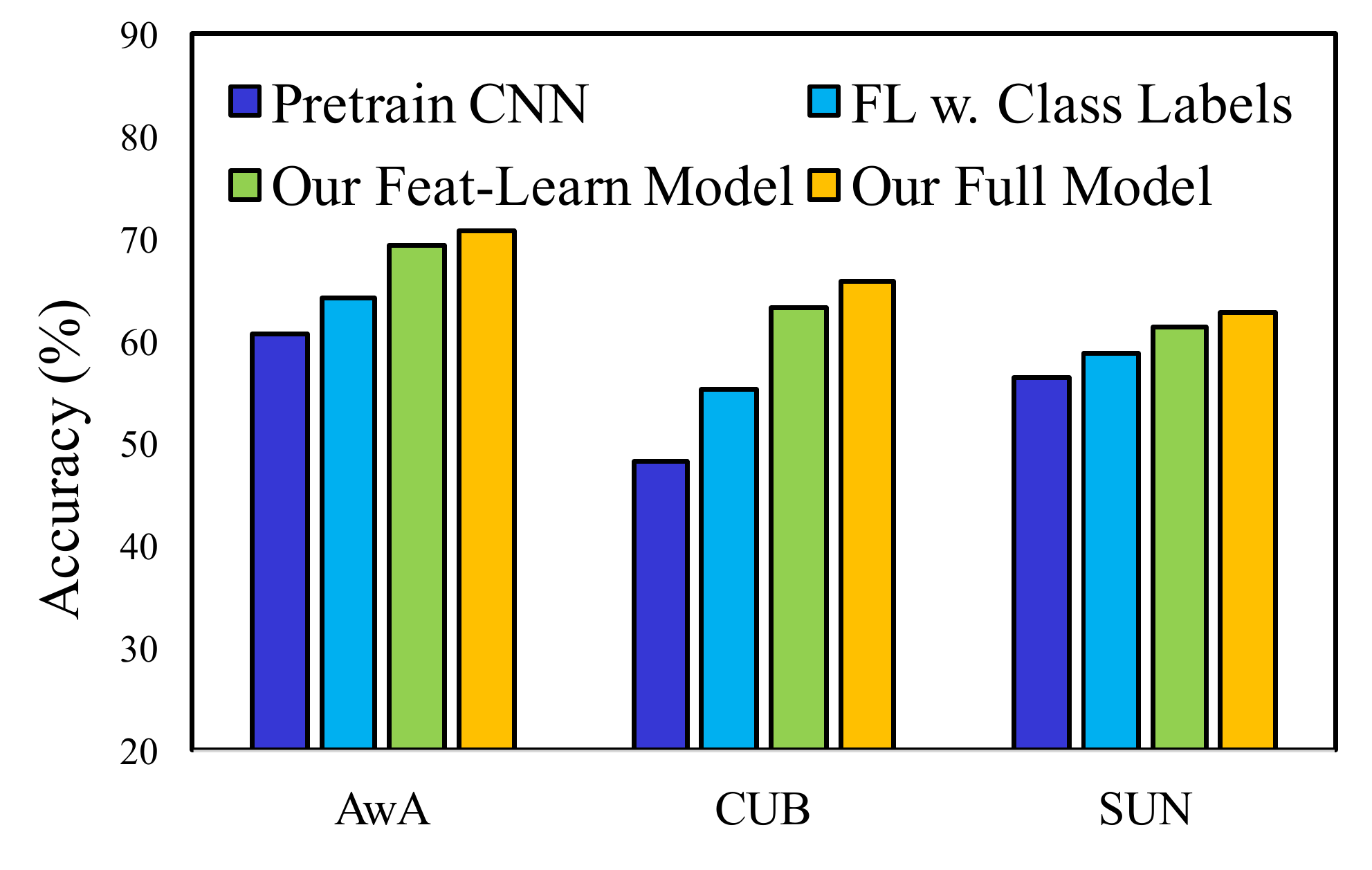}
\end{center}
\vspace{-0.1in}
\caption{Ablation study results on the three medium-scale datasets under the pure ZSL setting.}
\label{res_ab}
\end{figure}

\noindent\textbf{Generalised  ZSL}. We follow the same generalised ZSL setting of  \citep{Xian2017CVPR}. Table \ref{res_gz} provides the comparative results of generalised  ZSL on the three medium-scale datasets. We can observe that: (1) Different ZSL models take a different trade-off between $acc_u$ and $acc_s$, and the overall performance is thus measured by the HM metric. (2) Our model achieves similar results on seen and unseen class domains while existing models favor one over the other. That is, our model has the strongest generalisation ability under this more challenging setting. (3) Our model clearly yields the best overall performance on AwA and CUB datasets, while is outperformed by the state-of-the-art model \citep{Xian2018cvpr} on the SUN dataset. This is still very impressive, given that \citep{Xian2018cvpr} exploits a much superior CNN model (i.e. ResNet-101 \citep{he2016cvpr}) for feature extraction.

\begin{figure*}[t]
\vspace{0.02in}
\begin{center}
\begin{minipage}{0.28\linewidth}
\centerline	{\includegraphics[width=0.98\textwidth]{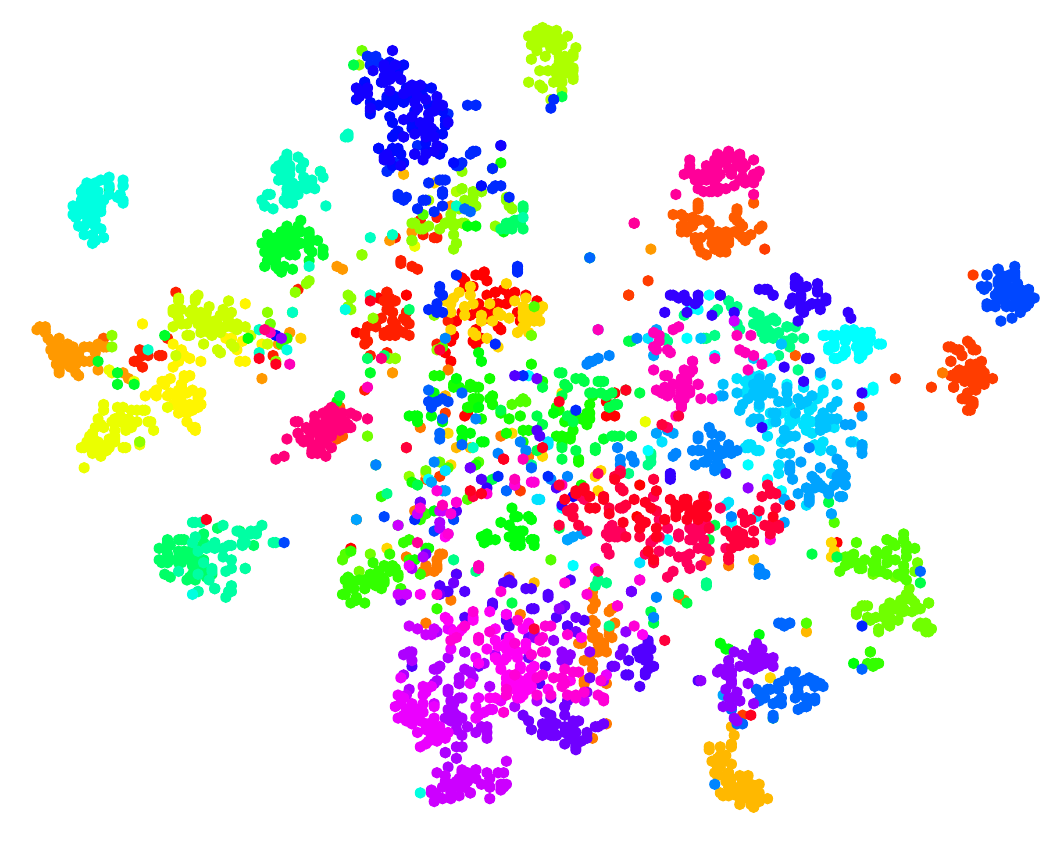}}\vspace{0.02in}
\centerline {(a) Pretrain CNN}
\end{minipage}
\hspace{0.08in}
\begin{minipage}{0.28\linewidth}
\centerline	{\includegraphics[width=0.98\textwidth]{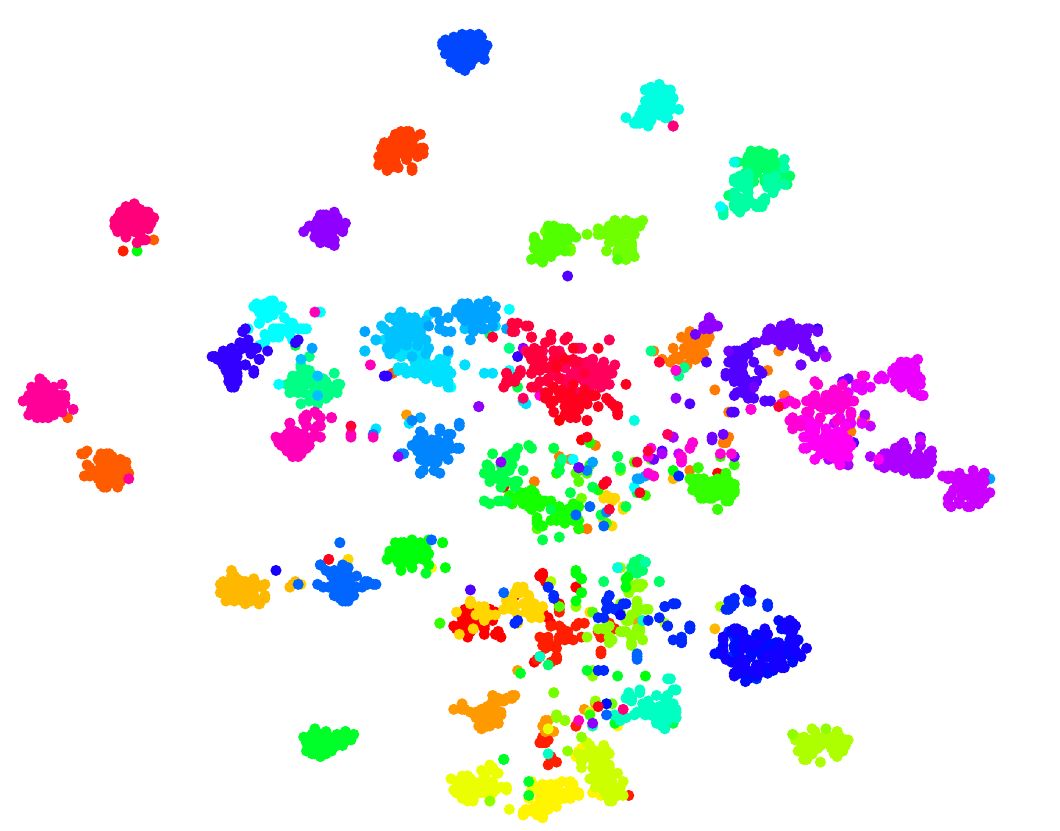}}\vspace{0.02in}
\centerline {(b) FL w. Class Labels}
\end{minipage}
\hspace{0.08in}
\begin{minipage}{0.28\linewidth}
\centerline	{\includegraphics[width=0.98\textwidth]{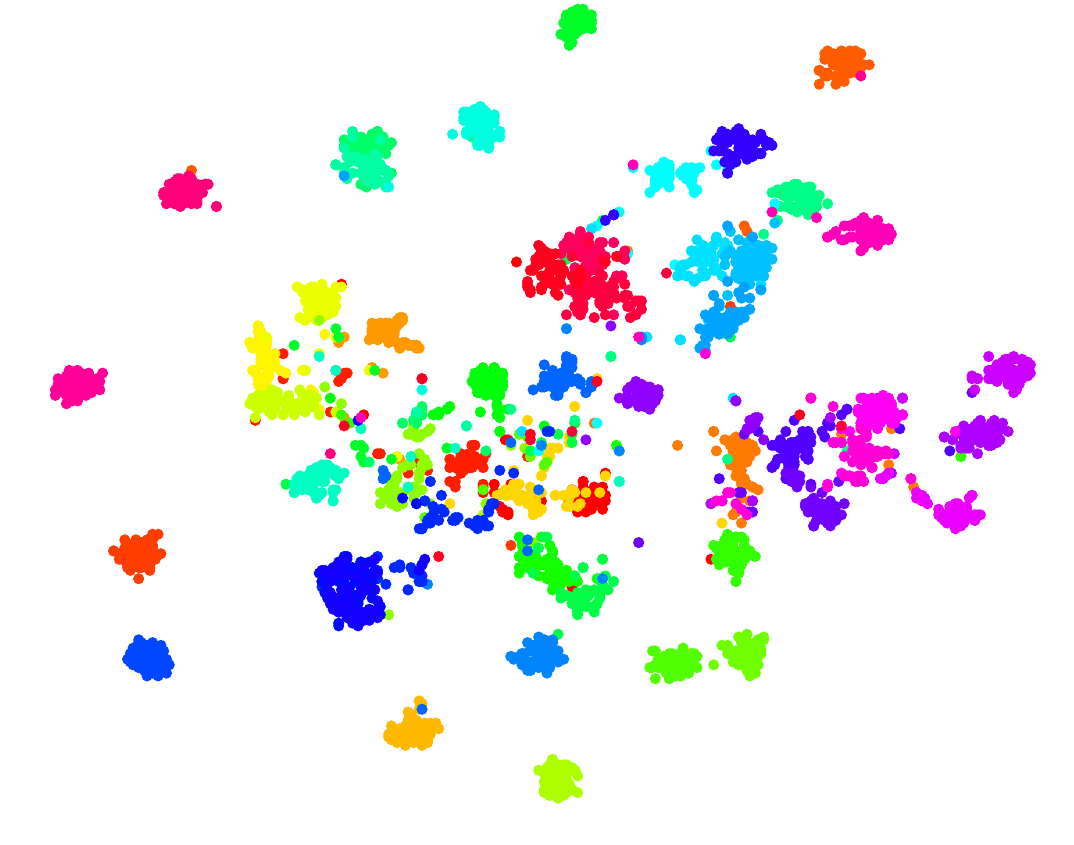}}\vspace{0.02in}
\centerline {(c) Our Feat-Learn Model}
\end{minipage}
\end{center}
\vspace{-0.15in}
\caption{The tSNE visualisation of the visual features of the test unseen class samples from the CUB dataset. The visual features of test samples are obtained by the feature learning methods used in the ablation study, while the labels (marked with different colors) of test samples are directly set as the ground-truth labels and thus keep fixed.}
\label{vis_cub}
\vspace{-0.02in}
\end{figure*}

\begin{figure*}[t]
\vspace{0.0in}
\begin{center}
\includegraphics[width=0.86\textwidth]{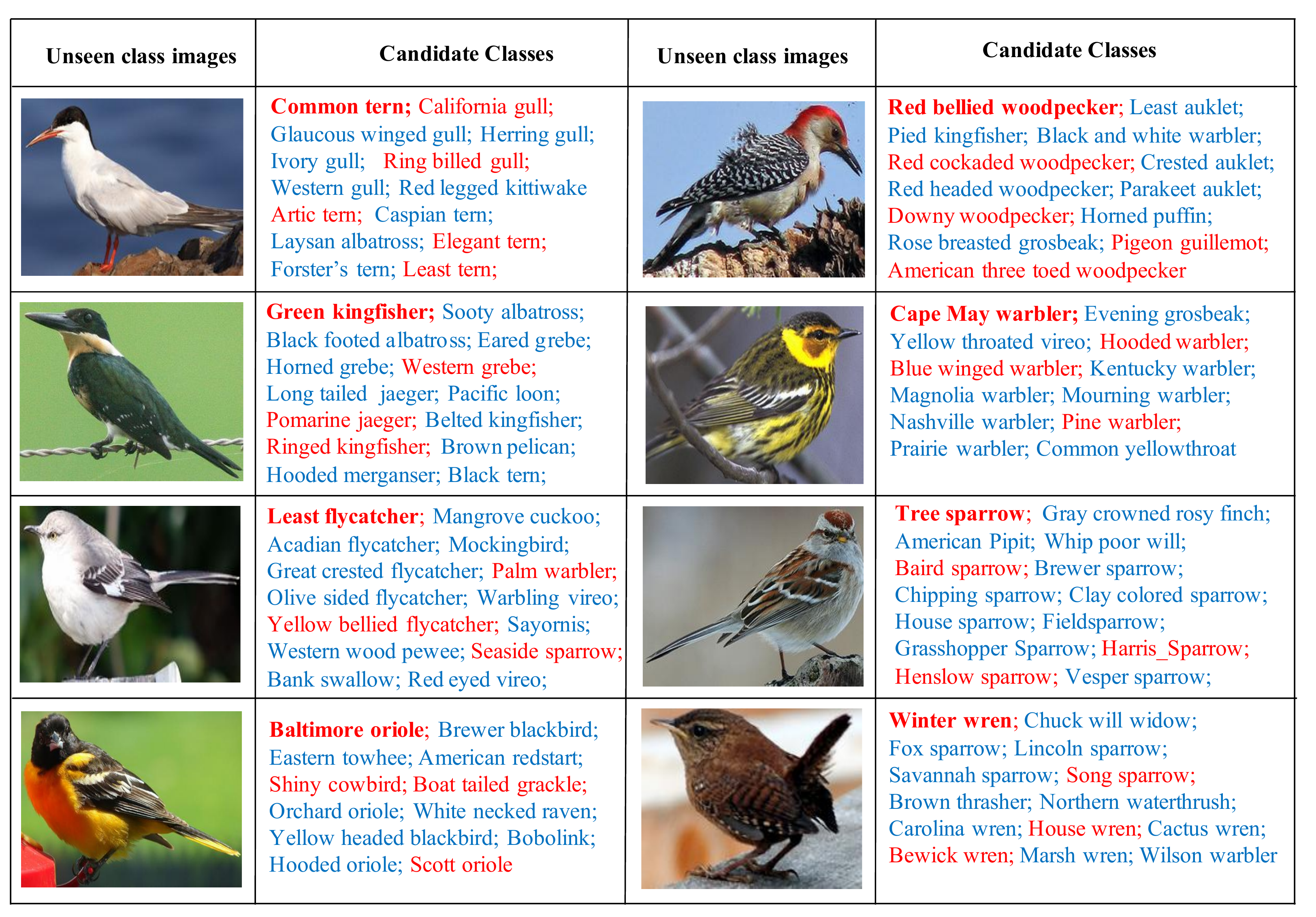}
\end{center}
\vspace{-0.15in}
\caption{Examples of candidate classes obtained by our ZSL model. The seen class candidates are in blue, unseen class candidates are in red, and the ground truth labels are in red bold. The number of candidate classes for each unseen class sample is forced to become smaller by projection function learning with superclasses.}
\vspace{-0.02in}
\label{vis_example}
\end{figure*}

\subsubsection{Further Evaluations}

\noindent\textbf{Ablation Study}. \label{sec:ab_study} We compare our full model with a number of striped-down versions to evaluate the effectiveness of the key components of our model. Three of such feature learning models are compared, each of which uses the same projection learning model (i.e. the baseline model defined by Eq.~(\ref{Eq:opt1}), where only class-level semantics are exploited) and differs only in how the visual features are obtained: `Pretrain CNN' -- the same CNN model trained on ImageNet without any finetuning on seen class data; `FL w. Class Labels' -- feature learning by finetuning with only class labels of seen class images; `Our Feat-Learn Model': feature learning by the proposed model in Sec.~\ref{sect:fea_learn}. The ablation study results in Fig.~\ref{res_ab} show that: (1) The transferrable feature learning with the class hierarchy leads to significant improvements (see Our Feat-Learn Model vs. FL w. Class Labels), ranging from 3\% to 9\%. This provides strong supports for our main contribution on deep feature learning for ZSL. (2) Our full model with extra transferrable projection learning achieves about 1--2\% improvements (see Our Full Model vs. Our Feat-Learn Model), showing the effectiveness of our transferrable projection learning for inductive ZSL. (3) As expected, finetuning the CNN model can learn better deep features for ZSL, yielding 2--7\% gains. However, the finetuned CNN model can not extract transferrable features and thus has very limited practicality.

\noindent\textbf{Qualitative Results}. We provide qualitative results to show why adding more components into our feature learning model benefits ZSL in the above ablation study. Fig.~\ref{vis_cub} shows the tSNE visualisation of the visual features of test unseen class samples. The visual features of test unseen class samples are obtained by the feature learning methods used in the above ablation study, while the labels of test unseen class samples are directly set as the ground-truth labels and thus keep fixed. It can be clearly seen that the visual features of the test samples become more separable when more components are added into our feature learning model, enabling us to obtain better recognition results.

\begin{table}[t]
\vspace{0.0in}
\begin{center}
\tabcolsep0.3cm
\caption{Comparative h@5 classification accuracies obtained by using the human-annotated/our hierarchy on the ImNet dataset. }
\label{ha_ab_imnet}
\vspace{-0.05in}
\begin{small}
\begin{tabular}{l|c|c}
\hline
Model & Human-annotated & Ours \\
\hline
Our Feat-Learn Model & 27.8 & \bf28.0\\
Our Full Model & 28.3 & \bf28.6\\
\hline
\end{tabular}
\end{small}
\end{center}
\vspace{-0.05in}
\end{table}

\begin{table}[t]
\vspace{0.0in}
\begin{center}
\tabcolsep0.3cm
\caption{Comparative classification accuracies obtained by using the human-annotated/our hierarchy on the CUB dataset.}
\label{ha_ab_cub}
\vspace{-0.05in}
\begin{small}
\begin{tabular}{l|c|c}
\hline
Model & Human-annotated & Ours \\
\hline
Our Feat-Learn Model & 64.6 & \bf64.7\\
Our Full Model & 65.6 & \bf65.8\\
\hline
\end{tabular}
\end{small}
\end{center}
\vspace{-0.05in}
\end{table}

In addition, some qualitative results are given to show how the proposed projection learning method benefits unseen class recognition. Fig.~\ref{vis_example} presents examples of candidate classes obtained by projection learning with superclasses. In this figure, the seen class candidates are in blue, unseen class candidates are in red, and the ground truth labels are in bold. According to Algorithm \ref{alg:pfl}, our projection learning method finds the best predicted label for each test image by nearest neighbor searching among unseen classes in several superclasses, rather than search among all unseen classes. This explains the superior performance of our projection learning method using superclasses.

\noindent\textbf{Human-Annotated Class Hierarchy}. The proposed ZSL model can be easily generalised to the human-annotated tree-structured class hierarchy (e.g. the biological taxonomy tree for animal classes, and the hierarchy tree of object classes provided by ImageNet). Concretely, we directly use seen/unseen classes as the bottom class layer and the higher-level classes are thus assigned as superclass layers (e.g. for animal classes, genus-level, family-level, and order-level layers can be used as superclass layers). Similarly, we can learn transferrable deep features by using the proposed feature learning model in Sec. \ref{sect:fea_learn}. With the learned features, we can infer the labels of test unseen class images by using the proposed projection learning method in Sec. \ref{sect:zsl_method}.

Tables \ref{ha_ab_imnet} and \ref{ha_ab_cub} give the comparative classification results with the human-annotated class hierarchy and our class hierarchy on the ImNet and CUB datasets under the pure ZSL setting, respectively. The human-annotated class hierarchy used in the CUB dataset is the biological taxonomy tree collected from Wikipedia, while the human-annotated class hierarchy used in the ImNet dataset is collected by WordNet\footnote{The class hierarchy is available at http://www.image-net.org.} \citep{wordnet}. As shown in these two tables, our class hierarchy has achieved better performance than the human-annotated class hierarchy. This is reasonable because the human-annotated class hierarchy is prone to the superclass imbalance (i.e. the number of classes belonging to different superclasses varies greatly), while our class hierarchy often provides more balanced clustering results.

\begin{table*}[t]
\vspace{0.00in}
\begin{center}
\tabcolsep0.5cm
\caption{Comparative 5-way accuracies (\%) with 95\% confidence intervals for FSL on mini-ImageNet.}
\label{fsl}
\vspace{-0.05in}
\begin{tabular}{lccc}
\hline\noalign{\smallskip}
Model & CNN& 1 shot & 5 shot \\
\noalign{\smallskip}\hline\noalign{\smallskip}
Nearest Neighbor&Simple&41.8$\pm$0.70&51.04$\pm$0.65\\
Matching Net \citep{vinyals2016bnips}&Simple&43.56$\pm$0.84&55.31$\pm$0.73\\
Meta-Learn LSTM\citep{Ravi2017iclr}&Simple&43.33$\pm$0.77&60.60$\pm$0.71\\
MAML\citep{Finn2017icml}&Simple&48.70$\pm$1.84&63.11$\pm$0.92\\
Prototypical Net \citep{Snell2017nips}&Simple&49.42$\pm$0.78&68.20$\pm$0.66\\
mAP-SSVM\citep{Triantafillou2017bnips}&Simple&50.32$\pm$0.80&63.94$\pm$0.72\\
Relation Net\citep{Sung2018cvpr}&Simple&50.44$\pm$0.82&65.32$\pm$0.70\\
SNAIL\citep{Mishra2018iclr}&ResNet20&55.71$\pm$0.99&68.88$\pm$0.92\\
PPA\citep{Qiao2018cvpr}&Simple&54.53$\pm$0.40&67.87$\pm$0.20\\
PPA\citep{Qiao2018cvpr}&WRN&\textbf{59.60$\pm$0.41}&73.74$\pm$0.19\\
\noalign{\smallskip}\hline\noalign{\smallskip}
Ours &Simple&53.92$\pm$0.63&65.13$\pm$0.61\\
Ours &WRN&57.14$\pm$0.78&\textbf{74.45$\pm$0.73}\\
\noalign{\smallskip}\hline
\end{tabular}
\end{center}
\vspace{-0.05in}
\end{table*}

\subsection{Few-Shot Learning}
\label{sect:res_fsl}

\subsubsection{Dataset and Settings}

To directly compare our method to the standard few-shot learning methods \citep{Snell2017nips, Qiao2018cvpr, Finn2017icml,Sung2018cvpr,Triantafillou2017bnips,Mishra2017,vinyals2016bnips}, we further apply it to FSL on mini-ImageNet as in \citep{Snell2017nips}. This dataset consists of 100 ImageNet classes, which is significantly smaller than the large-scale ImNet. The semantic space is formed as in Sec. \ref{zsl_exp}, while the visual features are extracted with two CNN models trained from scratch with the training set of mini-ImageNet: (1) Simple - four conventional blocks as in \citep{Snell2017nips}; (2) WRN - wide residual networks \citep{Zagoruyko2016bmvc} as in \citep{Qiao2018cvpr}. The 5-way accuracy is computed by randomly selecting 5 classes from the 20 novel classes for each test trial, and the average accuracy over 600 test trials is used as the evaluation metric.

\subsubsection{Comparative Evaluation}

The comparative 5-way accuracies on mini-ImageNet are given in Table \ref{fsl}. We can make the following observations: (1) Our model achieves state-of-the-art performance under the 5-way 5-shot setting and competitive results under 5-way 1-shot. This suggests that our model is effective not only for ZSL but also for FSL. (2) Stronger visual features extracted for FSL yield significantly better results, when PPA and our model are concerned.

\section{Conclusion}

In this paper, we tackle the domain gap challenge in ZSL by leveraging superclasses as the bridge between seen and unseen classes. We first build a tree-structured class hierarchy that consists several superclass layers and one single class layer. Then, we exploit the superclasses to overcome the domain gap challenge in two aspects: deep feature learning and projection function learning. In the deep feature learning phase, we introduce a recurrent neural network (RNN) defined with the superclasses from the class hierarchy into a convolutional neural network (CNN). A novel CNN-RNN model is thus proposed for transferrable deep feature learning. In the projection function learning phrase, we align the seen and unsee class domains using the superclasses from the class hierarchy. Extensive experiments on four benchmark datasets show that the proposed ZSL model yields significantly better results than the state-of-the-art alternatives. Furthermore, our model can be extended to few-shot learning and also achieves promising results. In our ongoing research, we would implement our current projection learning method using a deep autoencoder framework and then connect it to the CNN-RNN model so that the entire network can be trained in an end-to-end manner. Moreover, we also expect that our CNN-RNN model can be generalised to other ZSL-related vision problems such as multi-label ZSL and large-scale FSL.

\begin{acknowledgements}
This work was supported in part by National Natural Science Foundation of China (61573363 and 61573026), Beijing Natural Science Foundation (L172037), 973 Program of China (2015CB352502), the Fundamental Research Funds for the Central Universities and the Research Funds of Renmin University of China (15XNLQ01), and European Research Council FP7 Project SUNNY (313243).
\end{acknowledgements}

%\bibliographystyle{spbasic}  % basic style, author-year citations
%\bibliography{tfpl}   % name your BibTeX data base

\end{document}